\documentclass[10pt,twocolumn,letterpaper]{article}

\usepackage{cvpr}             
\usepackage[accsupp]{axessibility}
\usepackage{booktabs}
\usepackage{multirow}
\usepackage{MnSymbol}

\usepackage{float}

\definecolor{cvprblue}{rgb}{0.21,0.49,0.74}
\usepackage[pagebackref,breaklinks,colorlinks,allcolors=cvprblue]{hyperref}

\def\paperID{17} 
\def\confName{CVPR}
\def\confYear{2026}

\title{CrossFlowDG: Bridging the Modality Gap with Cross-modal Flow Matching for Domain Generalization}

\author{Antonios Kritikos\\
National Technical University of Athens\\
{\tt\small ankrit400@gmail.com}
\and
Nikolaos Spanos\\
National Technical University of Athens\\
{\tt\small nspanos@ails.ece.ntua.gr}
\and
Athanasios Voulodimos\\
National Technical University of Athens\\
{\tt\small thanosv@mail.ntua.gr}
}

\begin{document}
\maketitle
\begin{abstract}
Domain generalization (DG) aims to maintain performance under domain shift, which in computer vision appears primarily as stylistic variations that cause models to overfit to domain-specific appearance cues rather than class semantics. To overcome this, recent methods use textual representations as stable, domain-invariant anchors. However, multimodal approaches that rely on cosine similarity-based contrastive alignment leave a modality gap where image and text embeddings remain geometrically separated despite semantic correspondence. We propose \emph{CrossFlowDG}, a novel DG framework that addresses this residual gap using noise-free, cross-modal flow matching. By learning a continuous transformation in the joint Euclidean latent space, our framework explicitly transports domain-biased image embeddings toward domain-invariant text embeddings of the correct class.
Using the efficient VMamba image encoder and CLIP's text encoder, \emph{CrossFlowDG} is tested against four common DG benchmarks, and achieves competitive performance on several benchmarks and state-of-the-art on TerraIncognita.
Code is available at: \href{https://github.com/ajkrit/CrossFlowDG}{\texttt{https://github.com/ajkrit/CrossFlowDG}}
\end{abstract}    
\section{Introduction}

Domain Generalization (DG) studies a model’s ability to generalize to unseen target domains without access to target data during training~\cite{dgsurvey1, dgsurvey2}, a setting that reflects many practical deployment scenarios. In an era where machine learning systems are increasingly deployed across diverse real-world conditions, this capability is critical in applications such as autonomous driving across varying weather conditions and geographical regions~\cite{dgautonomous1, dgautonomous2, dgautonomous3}, and medical diagnosis systems that must operate reliably across different patient populations and imaging protocols~\cite{dgmedical1, dgmedical2, dgmedical3, dgmedical4}.

While traditional DG strategies rely on feature-level augmentations, adversarial alignment, or meta-learning, multimodal approaches have recently emerged as a powerful alternative~\cite{dgsurvey1, dgsurvey2}. By grounding visual features with language, these models aim to learn representations that are inherently more domain-invariant.


Despite the promise of using text as domain-invariant anchors in multimodal DG, standard cosine similarity-based contrastive alignment creates a \emph{modality gap}~\cite{mindthegap}. Because image and text embeddings occupy disjoint regions in the joint space, enforcing domain invariance by regularizing image features toward the text manifold becomes problematic. This observation raises an important question:


\begin{center}
     \emph{Does bridging the modality gap\\improve domain generalization?}
\end{center}

To address this question, we explore Flow Matching (FM)~\cite{fm}, a continuous normalizing flow framework that learns deterministic transport dynamics between probability distributions via ordinary differential equations. Conventional generative approaches, such as diffusion~\cite{songdiff, ddpm, sohl2015deep}, formulate text-to-image generation as a discrete denoising process, where the initial distribution is a simple Gaussian noise, and the text description is provided as an extra condition. On the contrary, the general formulation of FM allows for the construction of continuous mappings between arbitrary distributions.
In the multimodal DG setting, FM can be used to learn smooth transport maps that flow domain-biased image embeddings toward domain-invariant text embeddings of the same class.




The main contributions of this work are summarized as follows:

\begin{enumerate}
\item We introduce noise-free flow matching to explicitly bridge the modality gap between image and text embeddings.

\item We propose a DG framework, called \emph{CrossFlowDG}, that performs contrastive vision-language alignment followed by flow-based transport to map domain-biased image features toward domain-invariant text anchors, achieving state-of-the-art performance on the TerraIncognita dataset.

\end{enumerate}

\section{Related Work}

\paragraph{Domain Generalization.}
The field of DG has seen rapid evolution across several broad categories~\cite{dgsurvey1}. A common strategy is to explicitly align feature distributions across source domains through adversarial learning~\cite{domainal1, domainal2, domainal3} or statistical distance minimization~\cite{domainal4, domainal5}. Data augmentation methods instead synthesize novel domains to increase training diversity in image or feature space~\cite{dataaug1, dataaug2, dataaug3, dataaug4}. Optimization-based approaches promote generalization through worst-case objectives, loss landscape geometry, and contrastive regularization~\cite{groupdro, vrex, rsc, arm, swad, sagm, pcl}. A distinct line pursues causal or disentangled representations~\cite{mtl, idag, gmdg}. Transformer and state-space model backbones have also been explored as stronger inductive priors for DG~\cite{sdvit, gmoe, dgmamba, dgfamba}. Recently, SBGen~\cite{sbgen} proposed a Schrödinger Bridge~\cite{sb} framework to transport image features toward text anchors for DG using jointly pretrained CLIP encoders. In contrast, \emph{CrossFlowDG} employs noise-free flow matching between architecturally disparate encoders with no shared pretraining, demonstrating that explicit cross-modal transport generalizes beyond already-aligned embedding spaces.


\paragraph{Modality Gap.} Despite the success of vision-language models like CLIP~\cite{clip} in learning joint embeddings through cross-modal contrastive learning, a well-documented side effect of this training paradigm is the appearance of a \emph{modality gap}~\cite{mindthegap}: image and text embeddings occupy separate regions of the shared feature hypersphere. This phenomenon arises from a combination of architectural asymmetry between the visual and textual encoders and the nature of the InfoNCE objective, which optimizes relative rather than absolute alignment~\cite{mindthegap, alignclip}. Several strategies have been proposed to mitigate the gap, including post-hoc mean-centering~\cite{mindthegap}, parameter space sharing with intra-modality regularization~\cite{alignclip}, learned prior networks that explicitly translate between modal distributions as in DALL-E 2~\cite{dalle2}, and temperature scheduling to promote uniformity on the unit hypersphere~\cite{mitigatemodalitygap}.

\paragraph{Flow Matching.} Flow Matching (FM)~\cite{fm} has emerged as a powerful and flexible framework for learning continuous-time generative models by directly regressing a target vector field that transports a source distribution to a target distribution via an ODE. Building on the score-based perspective of diffusion models~\cite{ddpm, song2020score}, FM simplifies training by conditioning on pairs of source and target samples and adopting straight-line OT interpolants~\cite{stochasticinterpolants2022, stochasticinterpolants2023}, yielding a stable regression objective while retaining the theoretical guarantees of continuous normalizing flows~\cite{neuralode}. 
Noise-free FM variants, as in~\cite{crossflow}, where the source is a structured empirical distribution rather than Gaussian, have enabled deterministic cross-modal generation. 
FlowTok~\cite{flowtok} extends this by operating directly on 1D token representations across text and image modalities.

\section{Methodology}

\begin{figure*}[t]
    \centering
    \includegraphics[width=\linewidth]{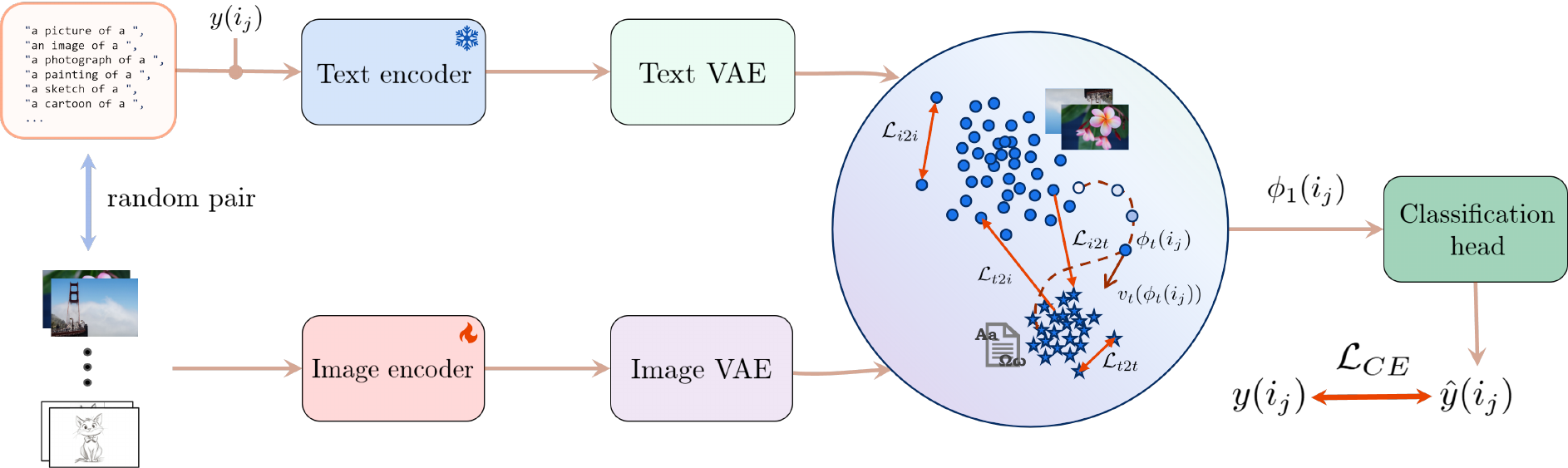}
    \caption{Overview of the \emph{CrossFlowDG} framework. Image samples are randomly paired with stylistic descriptions from the Textual Domain Bank (TDB). Then, the class corresponding to the image is appended to the textual description, and the unimodal inputs are fed to pretrained encoders. These intermediate representations are projected through VAEs in a joint latent space, which is formed via inter- and intra-modal cosine similarity-based contrastive alignment (Four-way Contrastive Loss, FCL). A Cross-modal Flow Matching (XFM) module tries to bridge the modality gap between embeddings of the same class.}
    \label{fig:method}
\end{figure*}

We propose \emph{CrossFlowDG}, a framework designed to explicitly address DG by bridging the gap between different modalities. The framework consists of three main components: (1) a Textual Domain Bank (TDB) that provides stylistically diverse semantic anchors, (2) a Four-way Contrastive Loss (FCL) that enforces intra- and inter-modal alignment, and (3) a Cross-modal Flow Matching (XFM) module that learns a deterministic mapping from domain-biased image representations to domain-invariant text representations. An overview of our method is illustrated in Fig. \ref{fig:method}.

\subsection{Textual Domain Bank (TDB)}
\label{sub:tdb}

A key challenge in DG is encouraging domain-invariant representations without requiring explicit domain labels or multiple annotated source domains. To address this, we introduce a Textual Domain Bank (TDB), which acts as a dynamic (but limited) prompt generator to supply stylistically diverse target representations.


We define a set of $k$ domain descriptors $\mathcal{D} = \{d_1, d_2, \dots, d_k\}$, where each $d_j$ represents a stylistic variation (\eg, ``\texttt{photograph}", ``\texttt{painting}", ``\texttt{drawing}", see Appendix~\ref{supp:tdb} for the complete list).

Before training, we pre-compute the embeddings for all prompt-class combinations to form a bank. More formally, let $c(\mathcal{I})$ denote the semantic class name corresponding to the set of same-class images $\mathcal{I}$. We uniformly sample a descriptor $d \sim \mathcal{U}(\mathcal{D})$ and construct the textual prompt $t$ via string concatenation ``$\oplus$": $$t(d, c(\mathcal{I})) = \text{``\texttt{a} "} \oplus d \oplus \text{``\texttt{of a} "} \oplus c(\mathcal{I}).$$
This process yields prompts such as ``\texttt{a photograph of a dog}", or ``\texttt{a sketch of a car}". 
During training, the prompt is sampled independently at each iteration, so the same image may be paired with different prompts across epochs.
As a result, the TDB creates a class-conditional distribution of text embeddings, which will serve as the target manifold for the subsequent alignment modules.


This pairing strategy serves two purposes. First, it acts as cross-modal data augmentation; by exposing each image to diverse stylistic framings, the model is forced to marginalize out the domain variable and extract purely domain-invariant semantic features. Second, it prevents representation collapse by inducing a distribution of text embeddings per class that establishes wider decision boundaries.

\subsection{Four-way Contrastive Loss (FCL)}

To organize the shared latent space and enforce domain invariance, we introduce a Four-way Contrastive Loss (FCL), which is composed of four complementary terms: (1) image-to-image ($\mathcal{L}_{i2i}$), (2) text-to-text ($\mathcal{L}_{t2t}$), and two cross-modal terms, image-to-text ($\mathcal{L}_{i2t}$) and text-to-image ($\mathcal{L}_{t2i}$).
The TDB supplies the FCL with semantically consistent but domain-mismatched hard positive pairs. This explicitly penalizes the network for relying on domain-specific shortcuts. The intra-modal losses ($\mathcal{L}_{i2i}$, $\mathcal{L}_{t2t}$) reduce intra-class variance, while the cross-modal losses ($\mathcal{L}_{i2t}$, $\mathcal{L}_{t2i}$) align the modality distributions of the same class.

Formally, let $f_i$ and $f_t$ denote pretrained image and text encoders, and $g_i$ and $g_t$ projection heads that project unimodal features into a shared latent space. The latent representations are:
\[
\mathbf{z}_j^i = g_i(f_i(\mathbf{x}_j)), \qquad
\mathbf{z}_j^t = g_t(f_t(\mathbf{t}_j)).
\]
For modalities $k$ and $l$ within a batch of size $N$, the InfoNCE~\cite{infonce} loss is:
\[
\mathcal{L}_{k2l} = -\frac{1}{N} \sum_{j=1}^{N}
\log
\frac{\exp(\text{cossim}(\mathbf{z}_j^k, \mathbf{z}_j^l) / \tau)}
{\sum_{m=1}^{N} \exp(\text{cossim}(\mathbf{z}_j^k, \mathbf{z}_m^l) / \tau)},
\]
where $\tau > 0$ is a temperature parameter, and $\text{cossim}(\cdot,\cdot)$ is the cosine similarity function.
The total contrastive objective is the balanced sum of all four pathways:
\[
\mathcal{L}_{\mathrm{FC}} =
\mathcal{L}_{i2t} + \mathcal{L}_{t2i}
+ \frac{1}{2}\left(\mathcal{L}_{t2t} + \mathcal{L}_{i2i}\right).
\]

\subsection{Cross-modal Flow Matching (XFM)}

Although FCL organizes the latent space, it fails to fully eliminate the geometric separation between images and text,
a phenomenon known as the \emph{modality gap}. In order to bridge the gap, we propose a noise-free, Cross-modal Flow Matching (XFM) module that learns a continuous transformation from image latents to text latents in the shared Euclidean space.

Given paired latent representations $(\mathbf{z}_{\text{img}}, \mathbf{z}_{\text{txt}})$, we define a linear interpolation at time $t \in [0,1]$:
\[
\mathbf{z}_t = (1 - t) \mathbf{z}_{\text{img}} + t \mathbf{z}_{\text{txt}}.
\]
The ground-truth velocity field governing this path is:
\[
\mathbf{v}^*(\mathbf{z}_t, t) = \frac{d \mathbf{z}_t}{dt} = \mathbf{z}_{\text{txt}} - \mathbf{z}_{\text{img}}.
\]
A neural network parameterized by $\theta$ is trained to predict a velocity field $\mathbf{v}_\theta(\mathbf{z}_t, t)$ by minimizing the mean squared error against the target vector:
\[
\mathcal{L}_{\mathrm{FM}}(\theta)
=
\mathbb{E}_{t \sim \mathcal{U}[0,1], (\mathbf{z}_{\text{img}}, \mathbf{z}_{\text{txt}}) \sim \pi_{\text{rand}}}
\left[
\| \mathbf{v}_\theta(\mathbf{z}_t, t) - \mathbf{v}^*(\mathbf{z}_t, t) \|_2^2
\right],
\]
where $\pi_{\text{rand}}$ denotes the random mini-batch coupling induced by the TDB sampling strategy.
Consequently, the XFM objective models a continuous vector field towards the class-conditional distribution shaped by the FCL.

During sampling, the learned flow defines a deterministic, modality-bridging mapping from image to text latent by simulating the ODE:
\[
\mathbf{z}_1 = \mathbf{z}_{\text{img}} + \int_0^1 \mathbf{v}_\theta(\mathbf{z}_t, t) \, dt.
\]
This flowed representation $\mathbf{z}_1$ is then used for classification.

\section{Experiments}

\subsection{Evaluation Metrics and Protocol}

We evaluate classification accuracy on four standard DG benchmarks: TerraIncognita~\cite{terraincognita}, PACS~\cite{pacs}, VLCS~\cite{vlcs}, and OfficeHome~\cite{officehome}. We adopt the leave-one-domain-out (LODO) evaluation protocol. Given a dataset with $N$ domains $\mathcal{D}_i$, $i \in \{1,2,\dots,N\}$, we train on $N-1$ domains and evaluate on the remaining one. This process is repeated for each domain as the test domain, and total performance is computed as the average:
\begin{equation}
\mathrm{LODO} = \frac{1}{N} \sum_{i=1}^{N} R_{\setminus i},
\label{eq:lodo}
\end{equation}
where $R_{\setminus i}$ denotes the classification accuracy when training on $\mathcal{D} \setminus \{\mathcal{D}_i\}$ and testing on $\mathcal{D}_i$.


\begin{table}[!t]
\centering
\caption{Performance comparison on TerraIncognita. \textbf{Bold} indicates best and \underline{underline} second-best.}
\label{tab:terra}
\small
\setlength{\tabcolsep}{3pt}
\resizebox{\columnwidth}{!}{%
\begin{tabular}{lccccccc}
\toprule
Method & Venue & Params & L100 & L38 & L43 & L46 & Avg. \\
\midrule
\multicolumn{8}{c}{\textit{ResNet-50 based}} \\
GroupDRO & ICLR'19 & 23M & 41.2 & 38.6 & 56.7 & 36.4 & 43.2 \\
VReX & ICML'21 & 23M & 48.2 & 41.7 & 56.8 & 38.7 & 46.4 \\
RSC & ECCV'20 & 23M & 50.2 & 39.2 & 56.3 & 40.8 & 46.6 \\
MTL & JMLR'21 & 23M & 49.3 & 39.6 & 55.6 & 37.8 & 45.6 \\
Mixstyle & ICLR'21 & 23M & 54.3 & 34.1 & 55.9 & 31.7 & 44.0 \\
SagNet & CVPR'21 & 23M & 53.0 & 43.0 & 57.9 & 40.4 & 48.6 \\
ARM & NeurIPS'21 & 23M & 49.3 & 38.3 & 55.8 & 38.7 & 45.5 \\
SWAD & NeurIPS'21 & 23M & 55.4 & 44.9 & 59.7 & 39.9 & 50.0 \\
PCL & CVPR'22 & 23M & 58.7 & 46.3 & 60.0 & 43.6 & 52.1 \\
SAGM & CVPR'23 & 23M & 54.8 & 41.4 & 57.7 & 41.3 & 48.8 \\
iDAG & ICCV'23 & 23M & 58.7 & 35.1 & 57.5 & 33.0 & 46.1 \\
GMDG & CVPR'24 & 23M & 59.8 & 45.3 & 57.1 & 38.2 & 50.1 \\
\midrule
\multicolumn{8}{c}{\textit{DeiT-S based}} \\
SDViT & ACCV'22 & 22M & 55.9 & 31.7 & 52.2 & 37.4 & 44.3 \\
GMoE & ICLR'23 & 34M & 59.2 & 34.0 & 50.7 & 38.5 & 45.6 \\
\midrule
\multicolumn{8}{c}{\textit{VMamba-T based}} \\
DGMamba & MM'24 & 31M & 62.0 & 47.7 & 61.7 & 46.9 & 54.5 \\
DGFamba & AAAI'25 & 31M & \underline{63.9} & \underline{49.8} & \textbf{63.1} & 47.5 & \underline{56.1} \\
\emph{CrossFlowDG} & 2026 & 36M & \textbf{66.6} & \textbf{52.0} & \underline{62.8} & \textbf{50.4} & \textbf{58.0} \\
\bottomrule
\end{tabular}%
}
\end{table}

\subsection{Datasets}

The \textbf{TerraIncognita} dataset contains 24{,}330 images collected from camera traps deployed across four different locations. The images depict animals from 10 distinct species and exhibit substantial environmental variation across domains.

\textbf{PACS} consists of 9{,}991 images spanning seven common object categories. The dataset is designed to capture large visual style shifts across domains, including photo, art painting, cartoon, and sketch.

\textbf{VLCS} includes 10{,}729 images collected from four distinct and popular image datasets. It provides a benchmark over five common object classes: \textit{bird}, \textit{car}, \textit{chair}, \textit{dog}, and \textit{person}.

\textbf{OfficeHome} comprises 15{,}588 images of 65 everyday object categories commonly found in office and home environments, such as \textit{Alarm-Clock}, \textit{Bed}, \textit{Chair}, and \textit{Mug}. The dataset features substantial domain diversity across different visual settings.


\subsection{Implementation and Experimental Setup}

We train \emph{CrossFlowDG} end-to-end using the following objective:
\begin{equation}
\mathcal{L}_{\text{total}} =
\lambda_{\text{img}} \mathcal{L}^{\text{img}}_{\text{VAE}} +
\lambda_{\text{txt}} \mathcal{L}^{\text{txt}}_{\text{VAE}} +
\lambda_{\text{FC}} \mathcal{L}_{\text{FC}} +
\lambda_{\text{FM}} \mathcal{L}_{\text{FM}} +
\lambda_{\text{CE}} \mathcal{L}_{\text{CE}},
\end{equation}
where $\mathcal{L}^{k}_{\text{VAE}} = \mathcal{L}^{k}_{\text{recon}} + \mathcal{L}^{k}_{\text{KL}}$ for modality $k \in \{\text{img}, \text{txt}\}$, with mean squared error (MSE) reconstruction and KL divergence regularization, $\mathcal{L}_{\text{CE}}$ is the standard cross-entropy loss between classifier logits and ground truth one-hot labels, and $\{ \lambda_{\text{img}}, \lambda_{\text{txt}}, \lambda_{\text{FC}}, \lambda_{\text{FM}}, \lambda_{\text{CE}} \}$ are empirically set to $\{ 0.3, 0.3, 0.3, 0.4, 10 \}$. The image encoder is the ImageNet-pretrained VMamba-T (29M parameters), whereas for text we use CLIP's text encoder, the only frozen component of our method. We also use two MLP-based VAEs (4M and 1M parameters), a ResNet-based flow model (1M parameters), and a MLP classification head (0.02M parameters), resulting in approximately 36M trainable parameters in total. The shared latent dimension is 256, and flow integration is performed using a 12-step Euler solver.

Training is performed for 10{,}000 iterations (50 epochs with 200 updates during each), with a batch size of 16 per source domain on a single NVIDIA A10G GPU, and takes approximately 1 minute per epoch. 

\section{Results}

Tables~\ref{tab:terra}--\ref{tab:officehome} summarize the performance of \emph{CrossFlowDG} across the four DG benchmarks. 

On \textbf{TerraIncognita} (Table~\ref{tab:terra}), \emph{CrossFlowDG} consistently improves over the previous state-of-the-art, achieving gains between 2.2\% and 3.5\% in three out of four target domains and yielding a +1.9\% increase in average accuracy. Both empirically (by simple inspection) and quantitatively (low accuracies compared to the other three datasets), TerraIncognita proves to be the most challenging of the evaluated benchmarks, due to variations in geographical location and environmental conditions, including low illumination, motion blur, and background variability. The observed improvement suggests that explicit cross-modal transport contributes to enhanced robustness under severe distribution shifts.

On \textbf{PACS} (Table~\ref{tab:pacs}), \emph{CrossFlowDG} achieves the highest accuracy in the \textit{Art} domain and the second-highest performance in the \textit{Cartoon} and \textit{Photo} domains, ranking second overall with a 0.5\% difference from the best-performing method. Performance in the \textit{Photo} domain is largely saturated across recent approaches using ImageNet pretrained backbones, with accuracies frequently exceeding 99\%, which limits the margin for measurable gains. 

On \textbf{VLCS} (Table~\ref{tab:vlcs}), \emph{CrossFlowDG} achieves state-of-the-art performance in one target domain, while its average accuracy remains 1.2\% below the leading method. Similar to PACS, certain VLCS domains also exhibit near-saturated performance.

Finally, on \textbf{OfficeHome} (Table~\ref{tab:officehome}), CrossFlowDG performs 2.9\% below the previous state-of-the-art on average. We attribute this degradation to the large label space of the dataset, which contains 65 object categories. Standard contrastive learning methods using the InfoNCE loss~\cite{infonce} require a sufficiently large batch size to provide a dense pool of negative samples~\cite{simclr}. In our experiments, the total batch size is 48 (16 per source domain), meaning at least 17 classes are entirely unrepresented in the denominator of the contrastive loss during any given forward pass. Consequently, the FCL objective is structurally constrained by a negative-sample deficit, preventing the formation of fully separated clusters for all 65 categories simultaneously. We hypothesize that a considerably larger effective batch size would improve performance.


\begin{table}[!t]
\centering
\caption{Performance comparison on PACS. \textbf{Bold} indicates best and \underline{underline} second-best.}
\label{tab:pacs}
\small
\setlength{\tabcolsep}{4pt}
\begin{tabular}{lccccc}
\toprule
Method & A & C & P & S & Avg. \\
\midrule
\multicolumn{6}{l}{\textit{ResNet-50 based}} \\
GroupDRO & 83.5 & 79.1 & 96.7 & 78.3 & 84.4 \\
VReX & 86.0 & 79.1 & 96.9 & 77.7 & 84.9 \\
RSC & 85.4 & 79.7 & 97.6 & 78.2 & 85.2 \\
MTL & 87.5 & 77.1 & 96.4 & 77.3 & 84.6 \\
Mixstyle & 86.8 & 79.0 & 96.6 & 78.5 & 85.2 \\
SagNet & 87.4 & 80.7 & 97.1 & 80.0 & 86.3 \\
ARM & 86.8 & 76.8 & 97.4 & 79.3 & 85.1 \\
SWAD & 89.3 & 83.4 & 97.3 & 82.5 & 88.1 \\
PCL & 90.2 & 83.9 & 98.1 & 82.6 & 88.7 \\
SAGM & 87.4 & 80.2 & 98.0 & 80.8 & 86.6 \\
iDAG & 90.8 & 83.7 & 98.0 & 82.7 & 88.8 \\
GMDG & 84.7 & 81.7 & 97.5 & 80.5 & 85.6 \\
\midrule
\multicolumn{6}{l}{\textit{DeiT-S based}} \\
SDViT & 87.6 & 82.4 & 98.0 & 77.2 & 86.3 \\
GMoE & 89.4 & 83.9 & 99.1 & 74.5 & 86.7 \\
\midrule
\multicolumn{6}{l}{\textit{VMamba-T based}} \\
DGMamba & 91.3 & 87.0 & 99.0 & \underline{87.3} & 91.2 \\
DGFamba & \underline{92.6} & \textbf{89.4} & \textbf{99.7} & \textbf{88.8} & \textbf{92.6} \\
\emph{CrossFlowDG} & \textbf{93.3} & \underline{89.0} & \underline{99.4} & 86.8 & \underline{92.1} \\
\bottomrule
\end{tabular}
\end{table}

\begin{table}[!t]
\centering
\caption{Performance comparison on VLCS. \textbf{Bold} indicates best and \underline{underline} second-best.}
\label{tab:vlcs}
\small
\setlength{\tabcolsep}{4pt}
\begin{tabular}{lccccc}
\toprule
Method & C & L & S & P & Avg. \\
\midrule
GroupDRO & 97.3 & 63.4 & 69.5 & 76.7 & 76.7 \\
VReX & 98.4 & 64.4 & 74.1 & 76.2 & 78.3 \\
RSC & 97.9 & 62.5 & 72.3 & 75.6 & 77.1 \\
MTL & 97.8 & 64.3 & 71.5 & 75.3 & 77.2 \\
Mixstyle & 98.6 & 64.5 & 72.6 & 75.7 & 77.9 \\
SagNet & 97.9 & 64.5 & 71.4 & 77.5 & 77.8 \\
ARM & 98.7 & 63.6 & 71.3 & 76.7 & 77.6 \\
SWAD & 98.8 & 63.3 & 75.3 & 79.2 & 79.1 \\
PCL & 99.0 & 63.6 & 73.8 & 75.6 & 78.0 \\
SAGM & 99.0 & 65.2 & 75.1 & 80.7 & 80.0 \\
iDAG & 98.1 & 62.7 & 69.9 & 77.1 & 76.9 \\
GMDG & 98.3 & {65.9} & 73.4 & 79.3 & 79.2 \\
\midrule
\multicolumn{6}{l}{\textit{DeiT-S based}} \\
SDViT & 96.8 & 64.2 & 76.2 & 78.5 & 78.9 \\
GMoE & 96.9 & 63.2 & 72.3 & 79.5 & 78.0 \\
\midrule
\multicolumn{6}{l}{\textit{VMamba-T based}} \\
DGMamba & 98.9 & 64.3 & 79.2 & \underline{80.8} & 80.8 \\
DGFamba & \textbf{99.5} & \textbf{66.2} & \underline{80.9} & \textbf{82.0} & \textbf{82.2} \\
\emph{CrossFlowDG} & 97.3 & \underline{66.1} & \textbf{81.3} & 79.5 & \underline{81.0} \\
\bottomrule
\end{tabular}
\end{table}

\begin{table}[!t]
\centering
\caption{Performance comparison on OfficeHome. \textbf{Bold} indicates best and \underline{underline} second-best.}
\label{tab:officehome}
\small
\setlength{\tabcolsep}{4pt}
\begin{tabular}{lccccc}
\toprule
Method & A & C & P & R & Avg. \\
\midrule
\multicolumn{6}{l}{\textit{ResNet-50 based}} \\
GroupDRO & 60.4 & 52.7 & 75.0 & 76.0 & 66.0 \\
VReX & 60.7 & 53.0 & 75.3 & 76.6 & 66.4 \\
RSC & 60.7 & 51.4 & 74.8 & 75.1 & 65.5 \\
MTL & 61.5 & 52.4 & 74.9 & 76.8 & 66.4 \\
Mixstyle & 51.1 & 53.2 & 68.2 & 69.2 & 60.4 \\
SagNet & 63.4 & 54.8 & 75.8 & 78.3 & 68.1 \\
ARM & 58.9 & 51.0 & 74.1 & 75.2 & 64.8 \\
SWAD & 66.1 & 57.7 & 78.4 & 80.2 & 70.6 \\
PCL & 67.3 & 59.9 & 78.7 & 80.7 & 71.6 \\
SAGM & 65.4 & 57.0 & 78.0 & 80.0 & 70.1 \\
iDAG & 68.2 & 57.9 & 79.7 & 81.4 & 71.8 \\
GMDG & 68.9 & 56.2 & 79.9 & 82.0 & 70.7 \\
\midrule
\multicolumn{6}{l}{\textit{DeiT-S based}} \\
SDViT & 68.3 & 56.3 & 79.5 & 81.8 & 71.5 \\
GMoE & 69.3 & 58.0 & 79.8 & 82.6 & 72.4 \\
\midrule
\multicolumn{6}{l}{\textit{VMamba-T based}} \\
DGMamba & \underline{76.2} & \underline{61.8} & \underline{83.9} & \underline{86.1} & \underline{77.0} \\
DGFamba & \textbf{77.4} & \textbf{63.7} & \textbf{85.6} & \textbf{87.3} & \textbf{78.5} \\
\emph{CrossFlowDG} & 74.9 & 60.9 & 82.6 & 84.0 & 75.6 \\
\bottomrule
\end{tabular}
\end{table}
\setcounter{topnumber}{2}
\setcounter{bottomnumber}{2}
\renewcommand{\floatpagefraction}{0.8}
\renewcommand{\textfraction}{0.05}

\section{Ablation Studies}

In order to empirically confirm the effect of each proposed component of \emph{CrossFlowDG} on the classification accuracy, we perform a series of ablation studies on the TerraIncognita dataset. In each experiment, we remove or restrict a single module and evaluate classification accuracy. The quantitative results are summarized in Table~\ref{tab:ablation}, demonstrating that the removal of any core component leads to a significant degradation in average accuracy.
Also, the effect of the number of ODE integration steps on the overall performance is investigated in Appendix~\ref{supp:ablation}.

\subsection{Ablation on TDB}
\label{sub:tdb}

We hypothesize that the TDB acts as a semantic regularizer by creating a distribution of same-class textual representations. 
To validate this, we restrict the TDB to generate only a single, static prompt template per class (\ie, ``\texttt{an image of a }"). As shown in Table~\ref{tab:ablation}, disabling the TDB's stylistic variance results in a severe performance drop of 5.1\% (from 58.0\% to 52.9\%). 

\subsection{Ablation on FCL}
\label{sub:fcl}

The FCL organizes the latent space by pulling intra-modal embeddings into compact clusters while pushing apart distinct classes. We hypothesize that this geometric structuring is a prerequisite for learning reliable flow trajectories. To test this, we remove the intra-modal terms of the FCL objective entirely. Without this contrastive geometric foundation, the cross-class manifolds (both at source and target) become scattered and overlapping, making the cross-modal transport harder to optimize. Consequently, average accuracy drops by 3.2\% (to 54.8\%), implying that the flow matching module relies on a well-structured latent space to prevent misalignment.

\subsection{Ablation on XFM}
\label{sub:xfm}

To demonstrate the superiority of continuous flow matching as a transport mechanism, we conduct an ablation where we keep the network architecture intact but replace the $\mathcal{L}_{\mathrm{FM}}$ objective with a standard Mean Squared Error (MSE) regression loss:
$$ \mathcal{L}_{\mathrm{MSE}} = \| \mathbf{z}_{\mathrm{txt}} - f_\theta(\mathbf{z}_{\mathrm{img}}) \|_2^2, $$
where $f_\theta(\cdot)$ is the mapping function (parameterized by the same ResNet used for our vector field in the main experiments), but now repurposed to directly predict the text latent coordinate.

As shown in Table~\ref{tab:ablation}, replacing the XFM module with MSE regression yields the most drastic degradation in the study, dropping the average accuracy by 5.6\% (to 52.4\%). 

\begin{table}[!t]
\centering
\caption{Ablation studies on TerraIncognita. ``---TDB" denotes ablation with a single-entry TDB, ``---FCL" denotes ablation keeping the inter-modal contrastive terms only, and ``---XFM" denotes ablation with simple ResNet mapping instead of the FM mechanism. \textbf{Bold} indicates best.}
\label{tab:ablation}
\small
\setlength{\tabcolsep}{4pt}
\begin{tabular}{lccccc}
\toprule
Method & L100 & L38 & L43 & L46 & Avg. \\
\midrule
\emph{CrossFlowDG} & \textbf{66.6} & \textbf{52.0} & {62.8} & \textbf{50.4} & \textbf{58.0} \\
\qquad --- TDB & {62.9} & {38.1} & \textbf{64.0} & {46.4} & 52.9 \\
\qquad --- FCL & {64.0} & {43.5} & {63.9} & {47.9} & 54.8 \\
\qquad --- XFM & {60.6} & {42.5} & {58.6} & {48.0} & 52.4 \\
\bottomrule
\end{tabular}
\end{table}

\subsection{Further Analysis}

To further understand the behavior of \emph{CrossFlowDG} and quantify the impact of our proposed components in greater detail, we conduct an analysis of the modality gap between image and text embeddings. Following prior work on multimodal representation learning, we evaluate several complementary metrics that capture different geometric aspects of the image-text embedding distributions.

To formalize these metrics, let $\mathcal{Z}_{c}^{\text{img}} = \{\mathbf{z}_{i,c}^{\text{img}}\}_{i=1}^{N_c}$ and $\mathcal{Z}_{c}^{\text{txt}} = \{\mathbf{z}_{j,c}^{\text{txt}}\}_{j=1}^{M_c}$ denote the sets of image and text embeddings for class $c$, with their respective centroids defined as $\mathbf{\mu}_{c}^{\text{img}} = \frac{1}{N_c}\sum_{i=1}^{N_c} \mathbf{z}_{i,c}^{\text{img}}$ and $\mathbf{\mu}_{c}^{\text{txt}} = \frac{1}{M_c}\sum_{j=1}^{M_c} \mathbf{z}_{j,c}^{\text{txt}}$.

\paragraph{Mean Absolute Modality Gap.}
Following along the lines of \cite{mindthegap}, we first measure the shift between modalities by computing the $\ell_2$ distance between the mean image embedding $\mu_{\text{img}}$ and the mean text embedding $\mu_{\text{txt}}$ for each class $c$, and then average them across the $C$ classes:
\begin{equation}
d_{\text{AMG}} = \frac{1}{C}{\sum_{c=1}^C\|\mathbf{\mu}_c^{\text{img}} - \mathbf{\mu}_c^{\text{txt}}\|_2} .
\end{equation}

\paragraph{Mean Relative Modality Gap.}
However, in order to be able to compare the modality gap across different experiments, we additionally report a normalized version of the $d_\text{AMG}$ measure. We define the dispersion $\sigma_c^k$ for modality $k \in \{\text{img}, \text{txt}\}$ as the mean Euclidean distance of its samples to the centroid: $\sigma_c^k = \frac{1}{|\mathcal{Z}_c^k|} \sum_{\mathbf{z} \in \mathcal{Z}_c^k} \|\mathbf{z} - \mathbf{\mu}_c^k\|_2$. The relative gap is then formulated as:
\begin{equation}
d_{\text{RMG}} = \frac{1}{C} \sum_{c=1}^C
\frac{\|\mu_c^{\text{img}} - \mu_c^{\text{txt}}\|_2}
{\sigma_c^{\text{img}} + \sigma_c^{\text{txt}}}.
\end{equation}

\paragraph{Mean Cosine Alignment.} While the $d_{\text{RMG}}$ metric controls for the size of the modality distributions, it does not rule out the possibility that the two distributions have collapsed towards the center of the Euclidean space with random orientation. Therefore, we report the mean cosine alignment, computed as follows:
\begin{equation}
    d_{\text{CA}} = \frac{1}{C} \sum_{c=1}^C \left( \frac{1}{N_c} \sum_{i=1}^{N_c} \text{cossim}(\mathbf{z}_{i,c}^{\text{img}}, \mathbf{\mu}_c^\text{txt}) \right),
\end{equation}
where $\mathbf{z}_{i,c}^{\text{img}}$ is replaced by the resulting flowed or mapped latent when evaluating the flow/mapping mechanism, and $\text{cossim}(\cdot, \cdot)$ is the cosine similarity function.



\begin{figure*}[t]
    \centering

    \begin{tabular}{@{} c @{\hspace{1em}} c @{\hspace{1em}} c @{}}
    (a) & \raisebox{-0.5\height}{\includegraphics[width=0.7\linewidth]{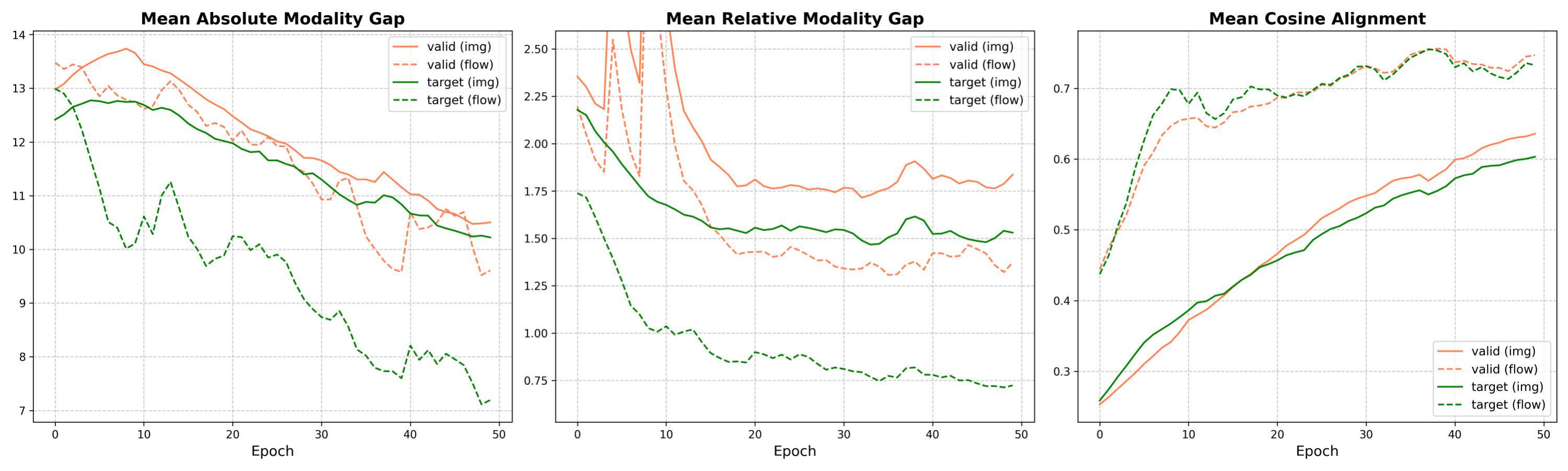}} 
        & \raisebox{-0.5\height}{\includegraphics[width=0.2\linewidth]{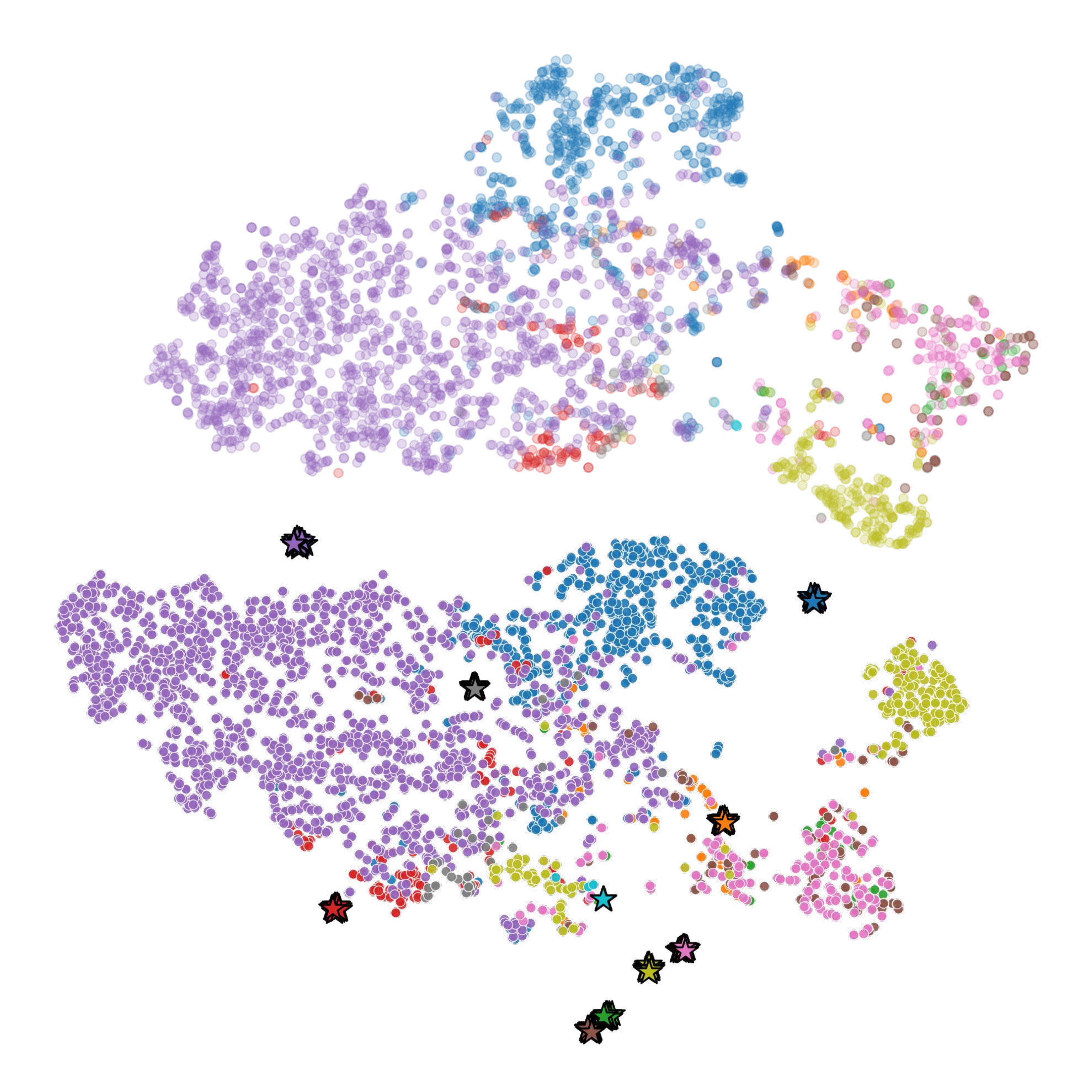}} \\
        [1.5ex] 
    (b) & \raisebox{-0.5\height}{\includegraphics[width=0.7\linewidth]{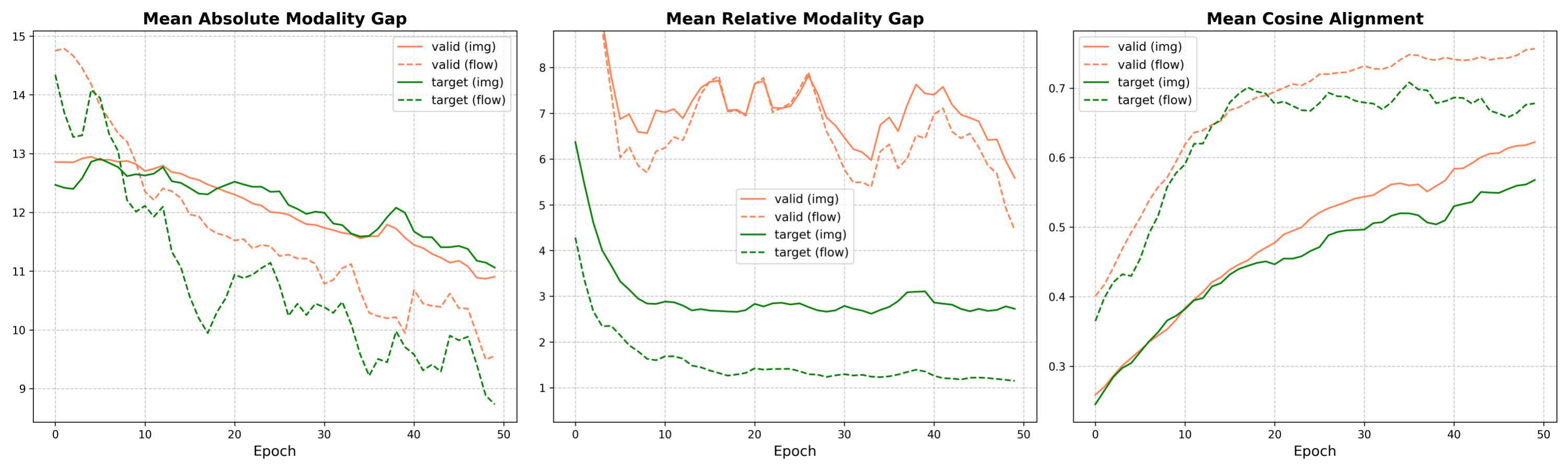}} 
        & \raisebox{-0.5\height}{\includegraphics[width=0.2\linewidth]{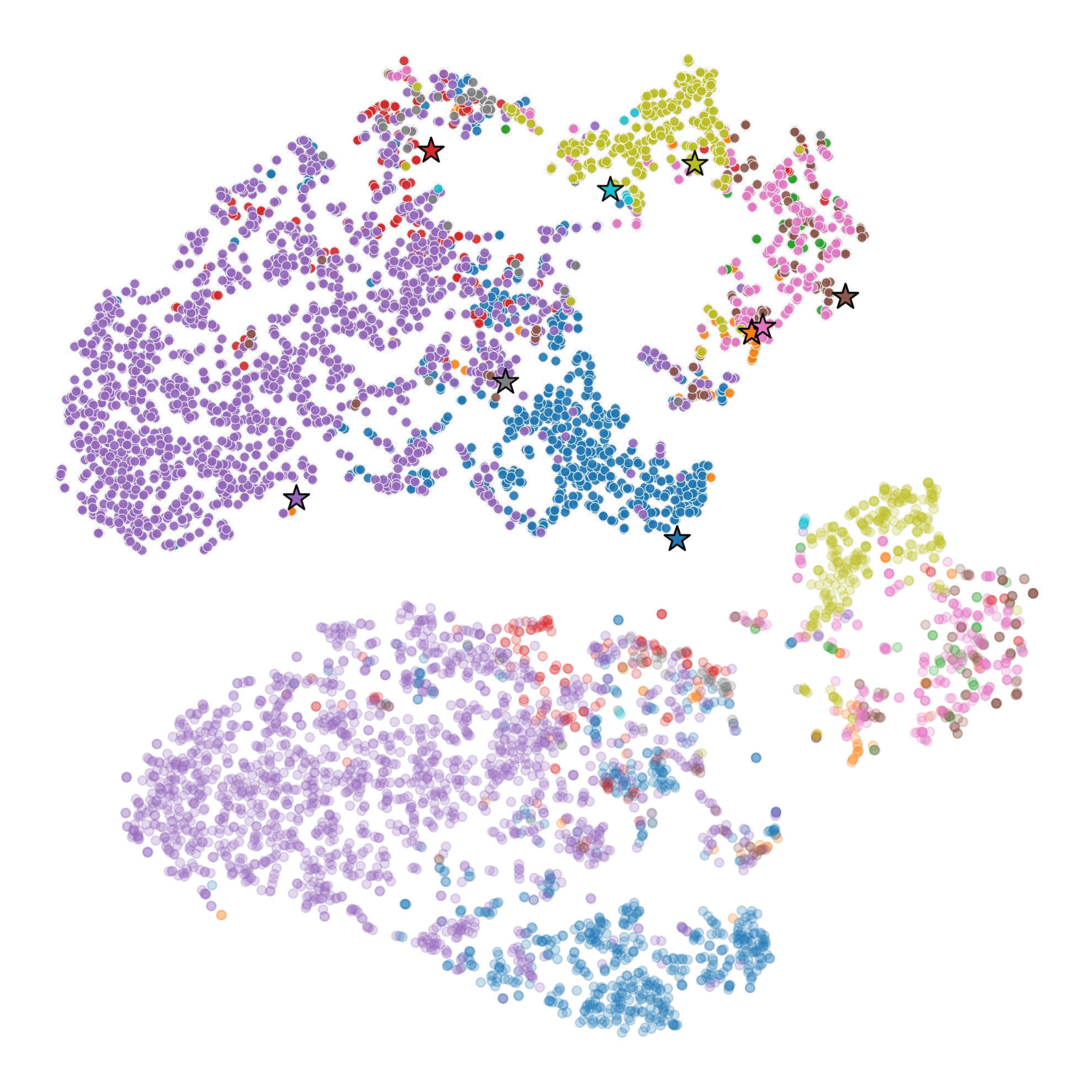}} \\
        [1.5ex]
    (c) & \raisebox{-0.5\height}{\includegraphics[width=0.7\linewidth]{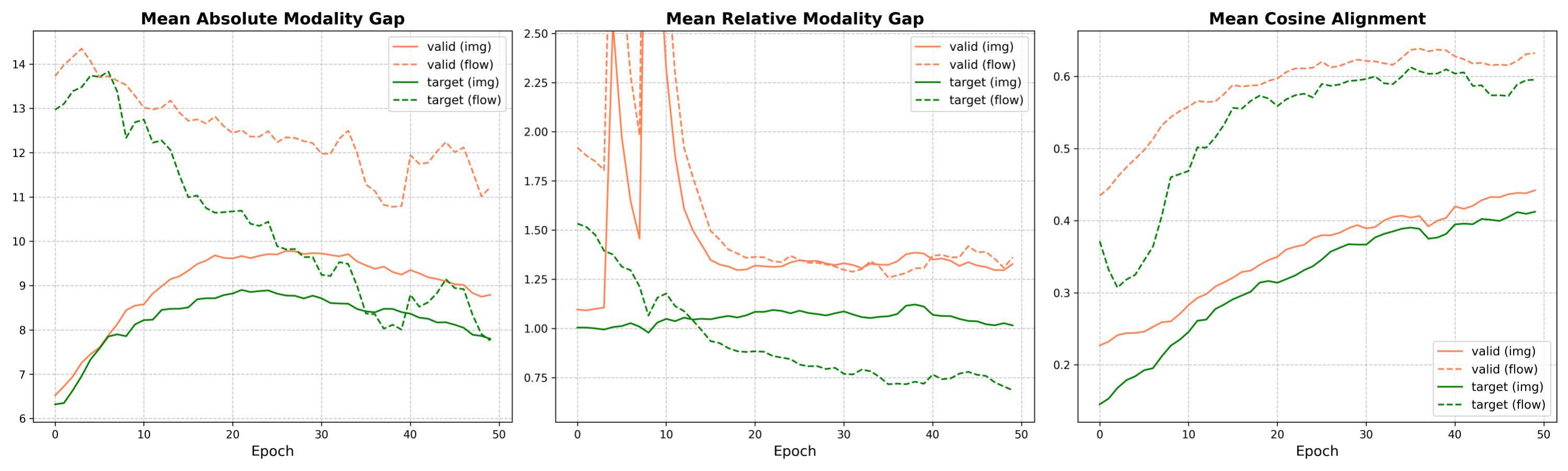}} 
        & \raisebox{-0.5\height}{\includegraphics[width=0.2\linewidth]{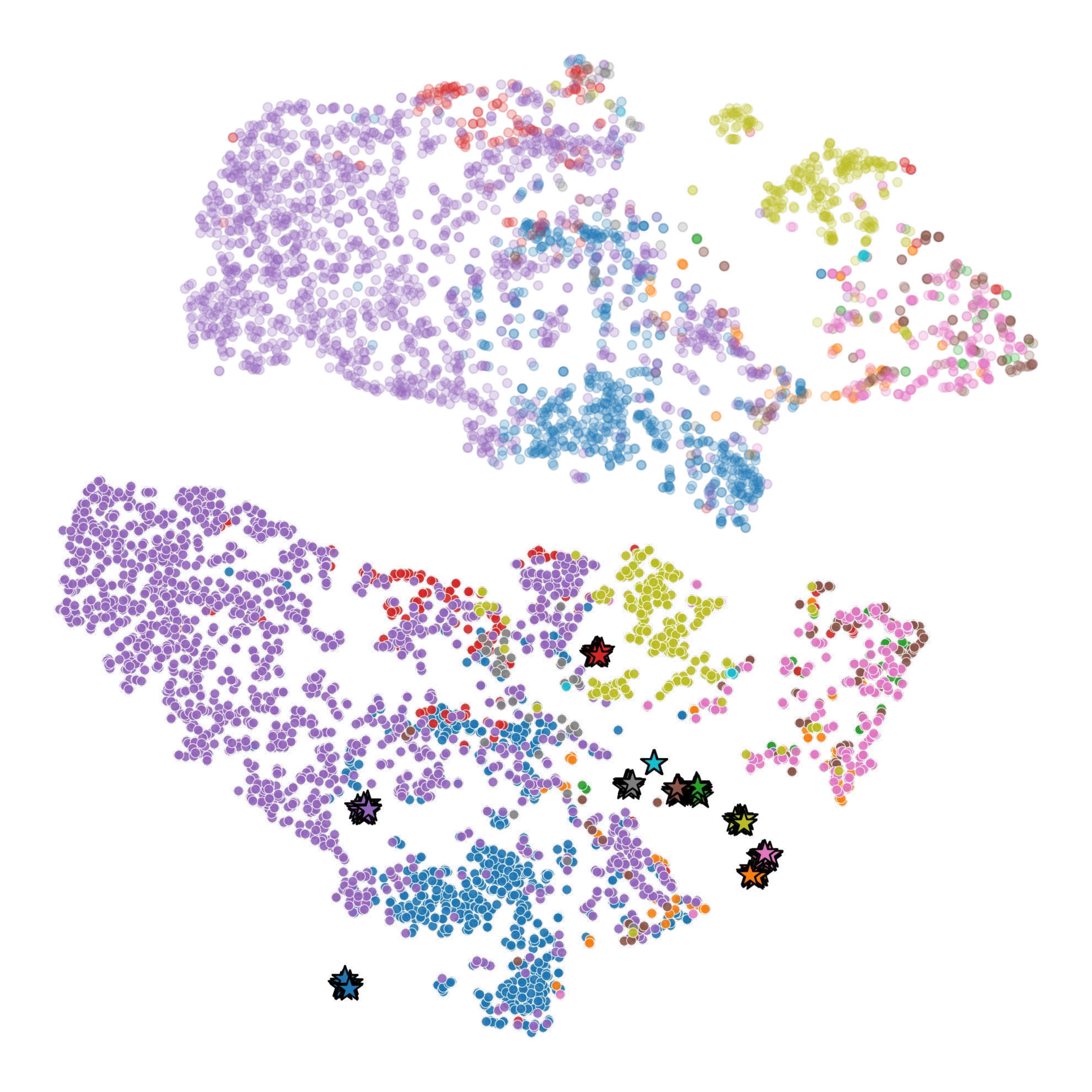}} \\
        [1.5ex]
    (d) & \raisebox{-0.5\height}{\includegraphics[width=0.7\linewidth]{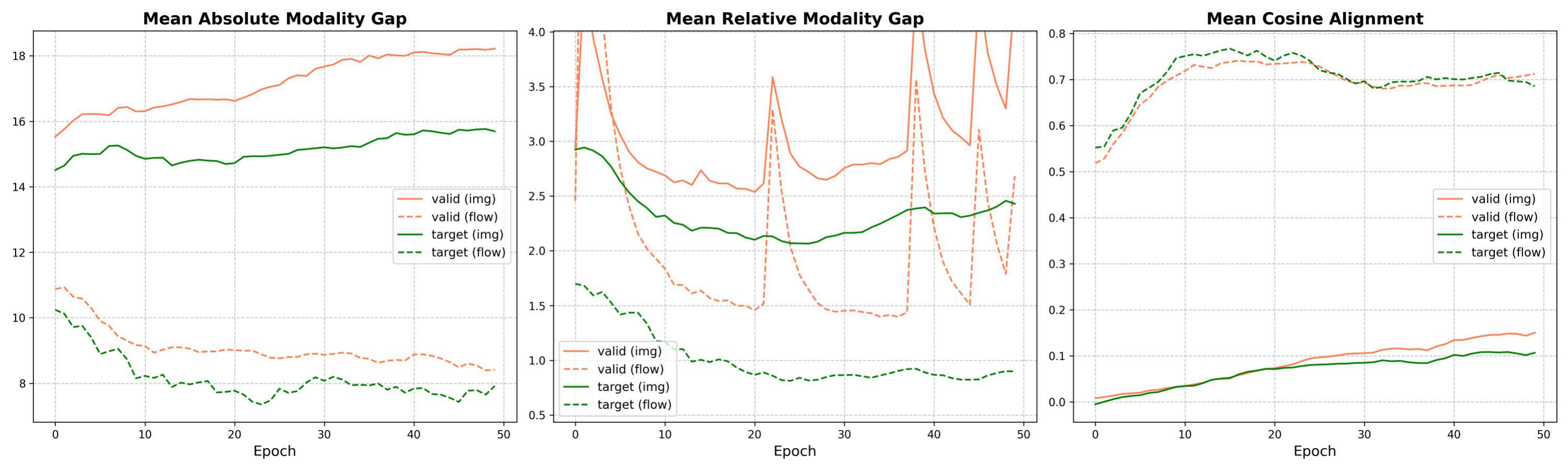}} 
        & \raisebox{-0.5\height}{\includegraphics[width=0.2\linewidth]{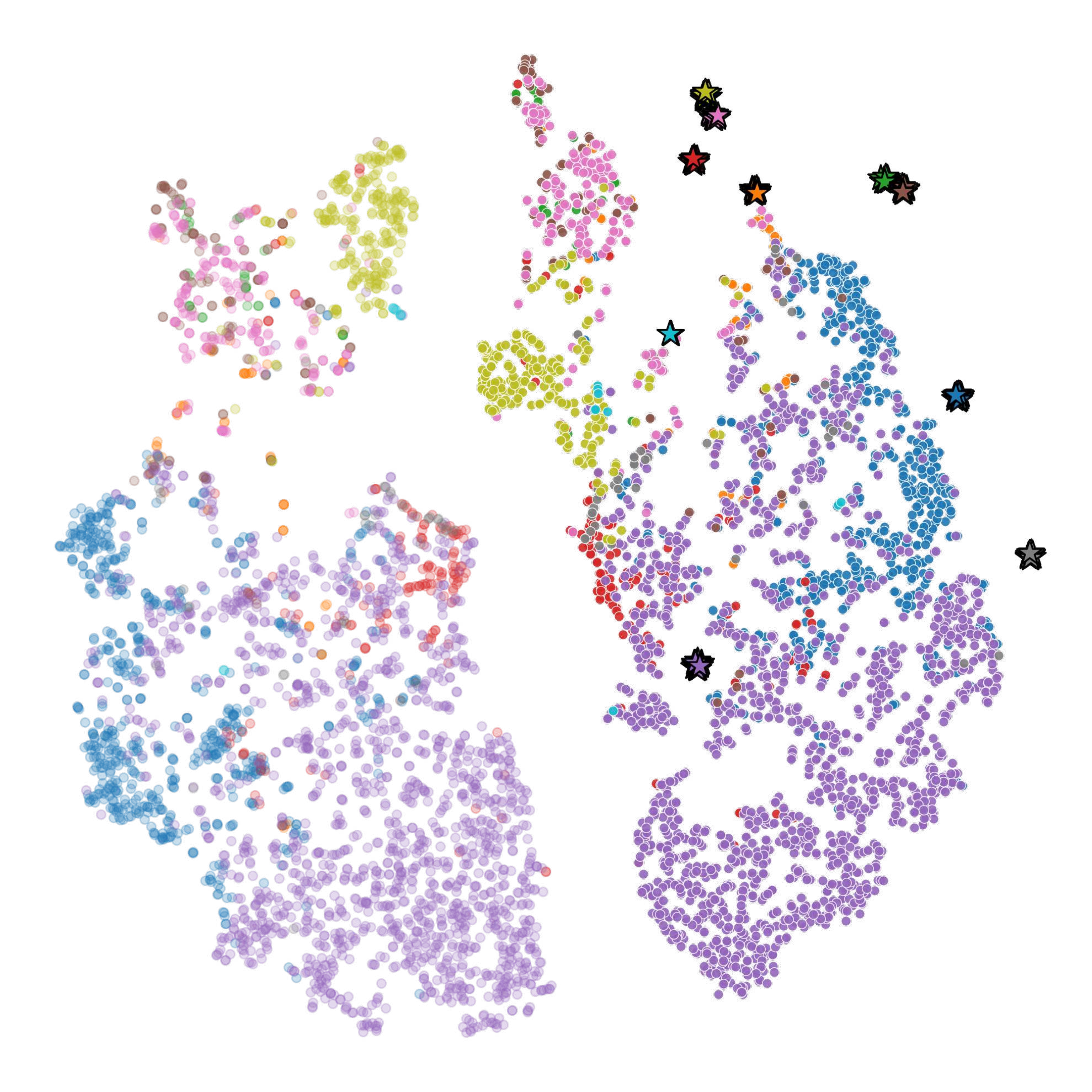}} \\
    (e) & \raisebox{-0.5\height}{\includegraphics[width=0.7\linewidth]{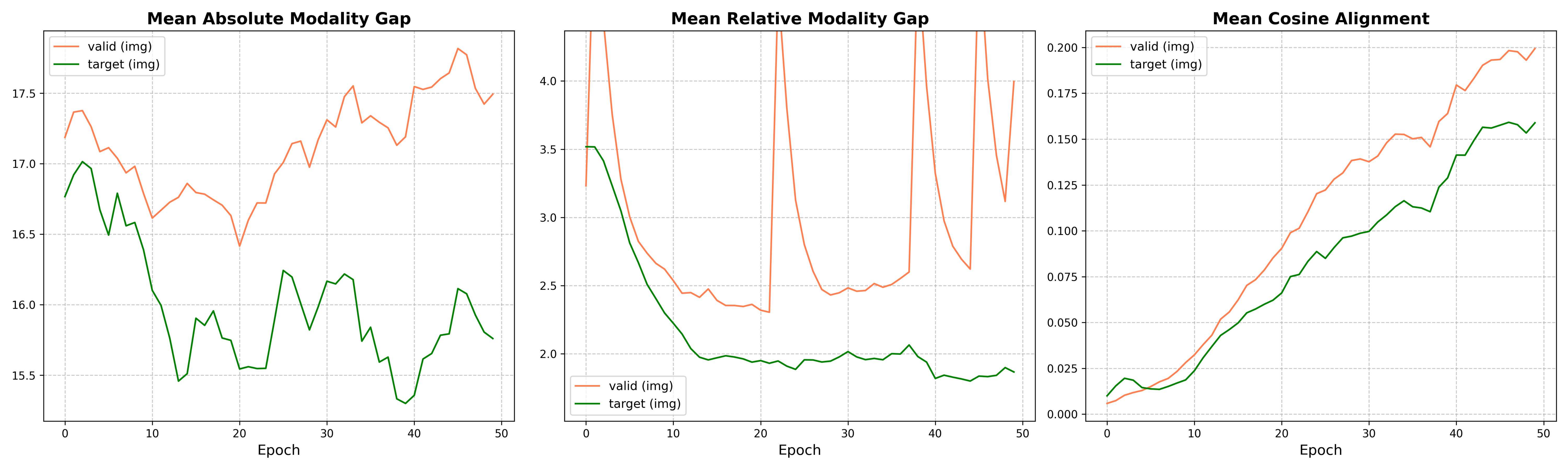}} 
        & \raisebox{-0.5\height}{\includegraphics[width=0.2\linewidth]{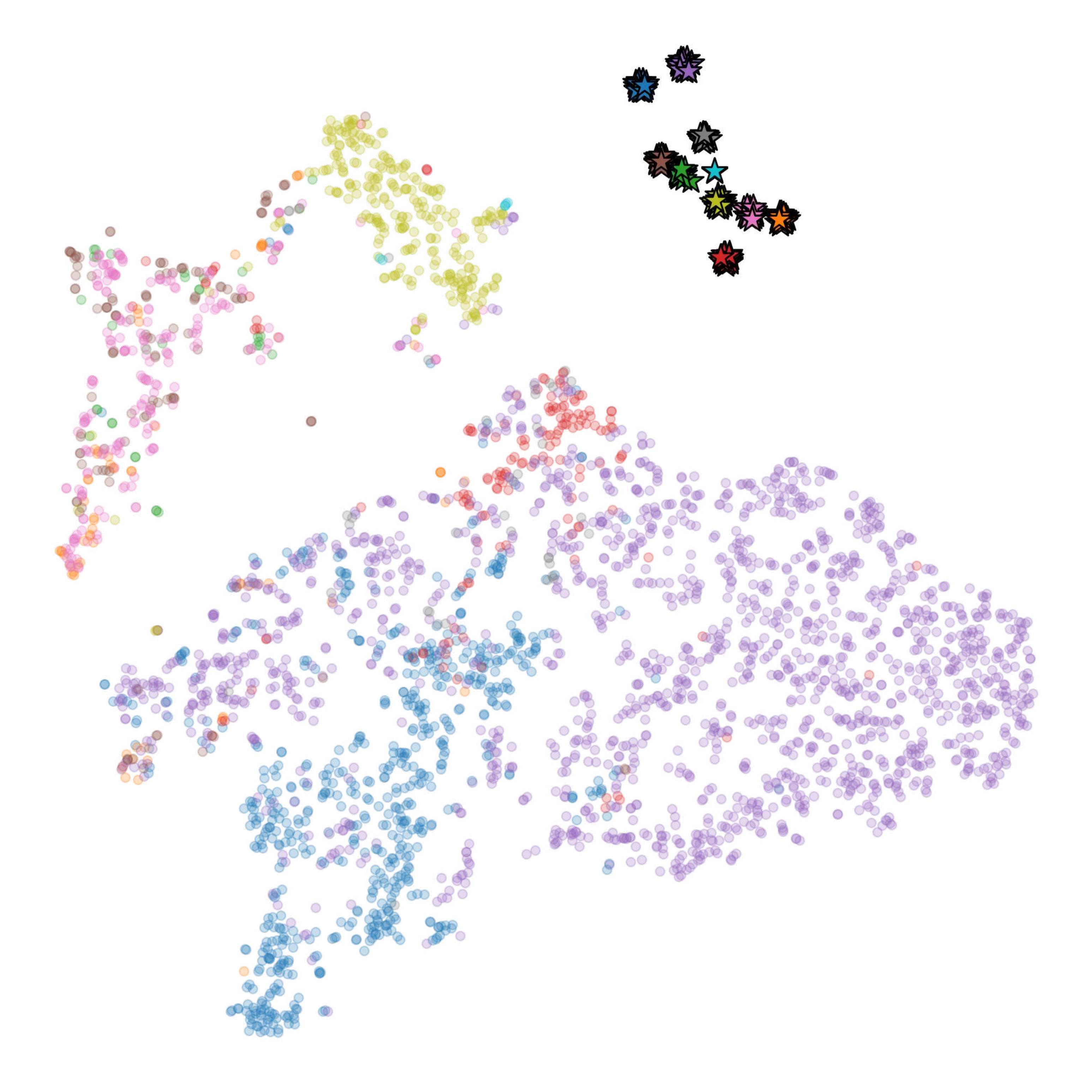}} \\
        
    \end{tabular}
    \caption{Plots of our proposed metrics ($d_{\text{AMG}}$, $d_{\text{RMG}}$, $d_{\text{CA}}$ in each row) and t-SNE visualization of the latent space of the L100 target domain of the TerraIncognita dataset across five experiments. \emph{Notes}: In the legend, ``img" denotes the initial image latent (as encoded by the VAE), while ``flow" denotes the resulting latent after flowing (or simple ResNet mapping in the case of (d)). For the $d_{\text{RMG}}$ metric, the plot is adjusted to show 95\% of the data for visual clarity. For the t-SNE visualization, the semi-transparent `$\bullet$' denote the image latents, the opaque `$\bullet$' denote the flowed (or mapped) image latents, and the `$\star$' denote the text latents. (a) \emph{CrossFlowDG}. (b) Ablation on TDB, as described in \ref{sub:tdb}. (c) Ablation on FCL, as described in \ref{sub:fcl}. (d) Ablation on XFM, as described in \ref{sub:xfm}. (e) Classification based on image latent representations, without any flow/mapping mechanism. The modality gap is evident.}
    \label{fig:metrics}
\end{figure*}

\paragraph{Observations.} The proposed metrics on L100, as well as t-SNE visualizations of the latent space across our experiments, are illustrated in Fig.~\ref{fig:metrics}. 

Our full proposed method (Fig.~\ref{fig:metrics}(a)) successfully bridges the modality gap for unseen domains. This is evidenced by the target flow achieving a significantly lower relative gap than the baseline images of the unseen domain. Furthermore, the cosine alignment of the target flow tightly tracks that of the validation flow, with both improving steadily over time. 

Removing key components of our architecture generates a noticeable generalization gap. In the ablations shown in Fig.~\ref{fig:metrics}(b) and Fig.~\ref{fig:metrics}(c), the target flow's cosine alignment explicitly diverges from the validation flow and while the validation flow continues to learn and align with the text anchors, the target flow struggles to follow suit. This indicates that the ablated models fail to generalize the cross-modal transport map to unseen distributions. 
Finally, Fig.~\ref{fig:metrics}(d) highlights the limitations of replacing the XFM module with a simple ResNet mapping optimized via a standard Mean Squared Error (MSE) objective. While the mapped flow representations manage to mimic the validation trajectory, the underlying raw image embeddings degrade severely. The absolute modality gap for the base images actually widens over time, and their cosine alignment is near zero. This indicates that the simple MSE mapping forces a superficial alignment at the output but actively degrades, or completely fails to structure, the underlying image distributions learned by the corresponding encoder.

In addition to quantitative metrics, we visualize the joint embedding space using t-SNE.
The right column of Fig.~\ref{fig:metrics} shows the distribution of image and text embeddings for the L100 domain of the TerraIncognita dataset. Under standard contrastive alignment and without using any of our proposed components, the two modalities occupy separate regions of the embedding space despite semantic correspondence, revealing the presence of a modality gap (Fig.~\ref{fig:metrics}(e)). \emph{CrossFlowDG} significantly reduces this separation by transporting image embeddings toward their corresponding text anchors, resulting in tighter cross-modal clusters (Fig.~\ref{fig:metrics}(a)).



\section{Discussion \& Future Work}

The proposed framework demonstrates that image-to-text transport can improve DG performance. Notably, our current architecture employs an asymmetric design: a VMamba-T visual backbone paired with a CLIP text encoder. Because these encoders are not jointly pretrained, our framework successfully forces alignment across fundamentally disparate latent spaces. This highlights the strength of the XFM module in bridging novel, unimodal visual backbones to rich language priors. Future work will investigate whether applying XFM to naturally symmetric vision-language models (\eg, dual CLIP encoders) yields similar relative gains, or if its primary utility lies in cross-architecture alignment.

Furthermore, while \emph{CrossFlowDG} establishes strong geometric alignment, it introduces trade-offs in scalability and inference speed. First, the Four-way Contrastive Loss (FCL) struggles with negative-sample deficits on datasets with large label spaces, like OfficeHome. Implementing momentum-based queues~\cite{moco} could decouple the negative sample pool from the hardware batch size, resolving this limitation. Second, simulating the ODE via a 12-step Euler integrator introduces latency during inference (see Appendix~\ref{supp:inference}). Future research will explore flow distillation or optimal transport mapping to achieve the exact same modality-bridging benefits with the speed of a single-step deterministic mapping, making the framework highly suitable for computationally constrained edge deployment.

Finally, several evaluated domains exhibit near-saturated performance, which may limit their ability to meaningfully differentiate generalization capabilities. Future work should therefore focus on constructing more challenging and diverse benchmarks that better reflect substantial real-world domain shifts, as well as the widespread use of pretrained backbones in modern applications.

\section{Conclusion}

This work presents \emph{CrossFlowDG}, a multimodal DG framework designed to explicitly bridge the residual modality gap left by standard contrastive alignment methods. Our framework comprises a Textual Domain Bank (TDB) and a Four-way Contrastive Loss (FCL) to establish a well-structured shared latent space and rich target manifolds. We then introduce noise-free, Cross-modal Flow Matching (XFM) to learn a continuous, deterministic transport function that maps domain-biased image representations directly onto these invariant text anchors in the shared Euclidean space. Extensive evaluations and ablations across four standard benchmarks suggest that bridging this modality gap improves robustness under domain shifts, allowing \emph{CrossFlowDG} to achieve state-of-the-art classification accuracy on the TerraIncognita dataset.

\section{Acknowledgements}
This work was supported by AWS resources, which were provided by the National Infrastructures for Research and Technology GRNET and funded by the EU Recovery and Resiliency Facility.
{
    \small
    \bibliographystyle{ieeenat_fullname}
    \bibliography{main}
}

\clearpage
\setcounter{page}{1}
\setcounter{section}{0}
\setcounter{table}{0}
\maketitlesupplementary

\renewcommand{\thesection}{\Alph{section}}
\renewcommand{\thetable}{S\arabic{table}}

\section{Textual Domain Bank Entries}
\label{supp:tdb}
Table~\ref{tab:prompts} enumerates the $k=18$ prompt templates used to construct the Textual Domain Bank (TDB), as described in Section~\ref{sub:tdb}.

\begin{table}[!ht]
\centering
\caption{Templates used in the Textual Domain Bank. 
Each template is completed with the target class name.}
\label{tab:prompts}
\small
\begin{tabular}{cl}
\toprule
\# & Template \\
\midrule
1  & a picture of a [class] \\
2  & an image of a [class] \\
3  & a photograph of a [class] \\
4  & a painting of a [class] \\
5  & a sketch of a [class] \\
6  & a cartoon of a [class] \\
7  & a 3D render of a [class] \\
8  & a drawing of a [class] \\
9  & a grayscale image of a [class] \\
10 & a low-light image of a [class] \\
11 & a high-resolution image of a [class] \\
12 & a blurred image of a [class] \\
13 & an overexposed image of a [class] \\
14 & a noisy image of a [class] \\
15 & a close-up image of a [class] \\
16 & a wide-angle image of a [class] \\
17 & an indoor image of a [class] \\
18 & an outdoor image of a [class] \\
\bottomrule
\end{tabular}
\end{table}


\section{Ablation on ODE Integration Steps}
\label{supp:ablation}
We investigate the trade-off between classification accuracy and inference efficiency by evaluating \emph{CrossFlowDG} across varying numbers of ODE integration steps, $N \in \{1, 6, 12\}$. The results are summarized in Table~\ref{tab:ablation_steps}.

\begin{table}[!ht]
\centering
\caption{Ablation on the number of ODE integration steps ($N$). \textbf{Bold} indicates best.}
\label{tab:ablation_steps}
\small
\setlength{\tabcolsep}{4pt}
\begin{tabular}{rccccc}
\toprule
$N$ & L100 & L38 & L43 & L46 & Avg. \\
\midrule
1 & {66.5} & {49.1} & {62.4} & {48.5} & {56.6} \\
6 & \textbf{67.2} & {48.1} & {62.5} & {47.9} & 56.4 \\
12 & {66.6} & \textbf{52.0} & \textbf{62.8} & \textbf{50.4} & \textbf{58.0} \\
\bottomrule
\end{tabular}
\end{table}

Notably, the single-step configuration ($N=1$) achieves an average accuracy of 56.6\%, namely a 1.4\% drop compared to the full 12-step model. It performs particularly well on the quantitatively easier domains (L100, L43), achieving accuracies comparable to the 12-step baseline. The 1-step performance empirically suggests that the learned vector field is highly consistent with the linear interpolation objective of flow matching, meaning a single Euler step provides a strong approximation of the true transport trajectory. While $N=6$ yields marginal improvements on the easier domains, it suffers a performance drop on the more challenging ones (L38, L46). The full $N=12$ configuration yields higher accuracy in three out of four domains, and the highest overall accuracy.
    
\section{Inference Efficiency}
\label{supp:inference}
Table~\ref{tab:flops} confirms that the lightweight flow model contributes minimally to the overall computational footprint; the dominant cost remains the VMamba-T backbone, which is executed exactly once per sample. Scaling the integration steps from $N=1$ to $N=12$ adds approximately 3.6 ms of overhead (11.47 ms vs. 15.13 ms on a consumer-grade NVIDIA RTX 4050 GPU). Because the flow map parameters ($\sim$1M) remain easily cached in VRAM, the iterative memory bottlenecks are largely bypassed. These baseline latency measurements on a consumer GPU indicate viability for resource-constrained edge deployment. Future work could explore flow distillation~\cite{boffi2025buildconsistencymodellearning, boffi2025flowmapmatchingstochastic} to achieve multi-step accuracy at the computational cost of a single-step forward pass.

\begin{table}[!ht]
\centering
\caption{Computational cost and inference latency across $N \in \{1, 6, 12\}$ ODE integration steps. Latency is measured as the average over 10,000 forward passes of a single sample on an NVIDIA RTX 4050 GPU.}
\label{tab:flops}
\small
\setlength{\tabcolsep}{4pt}
\begin{tabular}{rcc}
\toprule
$N$ & GFLOPs & Latency (ms) \\
\midrule
1 & 4.92 & 11.47 \\
6 & 4.98 & 13.22 \\
12 & 5.06 & 15.13 \\
\bottomrule
\end{tabular}
\end{table}

\end{document}